\documentclass[runningheads]{llncs}
\usepackage[T1]{fontenc}
\usepackage{graphicx,verbatim}
\usepackage[colorlinks=true, citecolor=blue, linkcolor=blue, urlcolor=blue]{hyperref}
\usepackage{color}

\usepackage{algorithm,algpseudocode}
\usepackage{multirow}
\usepackage{gensymb}
\usepackage{amsmath}
\usepackage{soul}
\usepackage{tikz}
\usepackage{tabularx}
\usepackage{orcidlink}
\usepackage{adjustbox}
\usepackage[skip=2pt]{caption}
\usetikzlibrary{calc}
\usepackage{hyperref}

\makeatletter
\newif\if@anonymize

\@anonymizefalse  

\if@anonymize
  \newcommand{\highlight@DoHighlight}{
    \fill [outer sep = -15pt, inner sep = 0pt, color=black]
          ($(begin highlight)+(0,8pt)$) rectangle ($(end highlight)+(0,-3pt)$) ;
  }

  \newcommand{\highlight@BeginHighlight}{
    \coordinate (begin highlight) at (0,0) ;
  }

  \newcommand{\highlight@EndHighlight}{
    \coordinate (end highlight) at (0,0) ;
  }

  \newdimen\highlight@previous
  \newdimen\highlight@current
  \newlength{\item@width}

  \DeclareRobustCommand*\anonymize{%
    \SOUL@setup
    \def\SOUL@preamble{%
      \begin{tikzpicture}[overlay, remember picture]
        \highlight@BeginHighlight
        \highlight@EndHighlight
      \end{tikzpicture}%
    }%
    \def\SOUL@postamble{%
      \begin{tikzpicture}[overlay, remember picture]
        \highlight@EndHighlight
        \highlight@DoHighlight
      \end{tikzpicture}%
    }%
    \def\SOUL@everyhyphen{%
      \discretionary{%
        \SOUL@setkern\SOUL@hyphkern
        \SOUL@sethyphenchar
        \tikz[overlay, remember picture] \highlight@EndHighlight ;%
      }{%
      }{%
        \SOUL@setkern\SOUL@charkern
      }%
    }%
    \def\SOUL@everyexhyphen##1{%
      \SOUL@setkern\SOUL@hyphkern
      \settowidth{\item@width}{##1}%
      \makebox[\item@width]{}%
      \discretionary{%
        \tikz[overlay, remember picture] \highlight@EndHighlight ;%
      }{%
      }{%
        \SOUL@setkern\SOUL@charkern
      }%
    }%
    \def\SOUL@everysyllable{%
      \begin{tikzpicture}[overlay, remember picture]
        \path let \p0 = (begin highlight), \p1 = (0,0) in \pgfextra
          \global\highlight@previous=\y0
          \global\highlight@current =\y1
        \endpgfextra (0,0) ;
        \ifdim\highlight@current < \highlight@previous
          \highlight@DoHighlight
          \highlight@BeginHighlight
        \fi
      \end{tikzpicture}%
      \settowidth{\item@width}{\the\SOUL@syllable}%
      \makebox[\item@width]{}%
      \tikz[overlay, remember picture] \highlight@EndHighlight ;%
    }%
    \SOUL@
  }
\else
  \newcommand{\anonymize}[1]{#1}
\fi
\makeatother

\begin{document}
\title{OccluNet: Spatio-Temporal Deep Learning for Occlusion Detection on DSA}
\titlerunning{OccluNet}

\author{Anushka A. Kore\inst{1}\orcidlink{0009-0003-5434-1396} \and Frank G. te Nijenhuis\inst{2}\orcidlink{0009-0003-1321-5836} \and Matthijs van der Sluijs\inst{2}\orcidlink{0000-0002-4934-0933} \and Wim van Zwam\inst{3}\orcidlink{0000-0003-1631-7056} \and Charles Majoie\inst{4}\orcidlink{0000-0002-7600-9568} \and Geert Lycklama {\`a} Nijeholt\inst{5}\orcidlink{0000-0001-6575-9868} \and Danny Ruijters\inst{6}\orcidlink{0000-0002-9931-4047} \and Frans Vos\inst{1}\orcidlink{ 0000-0003-2996-6872} \and Sandra Cornelissen\inst{2}\orcidlink{0000-0002-0332-2158} \and Ruisheng Su\inst{6}\orcidlink{0000-0002-5013-1370} \and Theo van Walsum\inst{2}\orcidlink{0000-0001-8257-7759}} 
\institute{
TU Delft, Mekelweg 5, 2628 CD Delft, The Netherlands\\
\email{anushkakore13@gmail.com, f.m.vos@tudelft.nl}
\and
Erasmus MC, Doctor Molewaterplein 40, 3015 GD Rotterdam, The Netherlands\\
\email{\{f.tenijenhuis,p.m.vandersluijs,s.cornelissen,t.vanwalsum\}@erasmusmc.nl}
\and
Maastricht UMC+, P. Debyelaan 25, 6229 HX Maastricht, The Netherlands\\
\email{w.van.zwam@mumc.nl}
\and
Amsterdam UMC, Meibergdreef 9, 1105 AZ Amsterdam, The Netherlands\\
\email{c.b.majoie@amsterdamumc.nl}
\and
Haaglanden MC, Lijnbaan 32, 2512 VA Den Haag, The Netherlands\\
\email{g.lycklama.a.nijeholt@haaglandenmc.nl}
\and
TU Eindhoven, Groene Loper 5, 5612 AE Eindhoven, The Netherlands\\
\email{\{d.ruijters,r.su\}@tue.nl}}
\authorrunning{A. A. Kore \textit{et al.}}











    
\maketitle              
\begin{abstract}
Accurate detection of vascular occlusions during endovascular thrombectomy (EVT) is critical in acute ischemic stroke (AIS). Interpretation of digital subtraction angiography (DSA) sequences poses challenges due to anatomical complexity and time constraints. This work proposes OccluNet, a spatio-temporal deep learning model that integrates YOLOX, a single-stage object detector, with transformer-based temporal attention mechanisms to automate occlusion detection in DSA sequences. We compared OccluNet with a YOLOv11 baseline trained on either individual DSA frames or minimum intensity projections. Two spatio-temporal variants were explored for OccluNet: pure temporal attention and divided space-time attention. Evaluation on DSA images from the \anonymize{MR CLEAN Registry} revealed the model's capability to capture temporally consistent features, achieving precision and recall of 89.02\% and 74.87\%, respectively. OccluNet significantly outperformed the baseline models, and both attention variants attained similar performance. 
Source code is available \href{https://github.com/anushka-kore/OccluNet.git}{here}. 


\end{abstract}

\section{Introduction}

Acute Ischemic Stroke (AIS) is a potentially life-threatening medical emergency caused by a sudden interruption of blood flow to a region of the brain due to a vessel occlusion \cite{IschemicStrokeBackground2025}. Endovascular thrombectomy (EVT) is a minimally invasive intervention to restore cerebral perfusion through mechanical removal of the occlusion. During EVT procedures, digital subtraction angiography (DSA) images are acquired to visualize the occlusion and to serve as a roadmap to navigate to the occlusion \cite{czapOverviewImagingModalities2021}. DSA is acquired in 2-4 frames per second to visualize the contrast agent flowing through the vasculature. It remains the gold standard for luminal imaging, delivering superior spatial and temporal resolution of the cerebral vasculature. 
Recent advances in deep learning (DL) have demonstrated potential in automating occlusion detection in various imaging modalities used for the diagnosis and management of stroke \cite{chavvaDeepLearningApplications2022}. For DSA imaging in particular, recent studies to automate occlusion detection relied on DL methods applied to static frames. Karlsberg et al. \cite{karlsbergPredictiveDeepLearning2023} employed YOLOv3 (You Only Look Once), a single-stage object detector, to detect occlusions on single-frame datasets derived from DSA sequences. Khankari et al. \cite{khankariAutomatedDetectionArterial2023a} utilized a ResNet-50 convolutional neural network (CNN) to analyze patch-sampling-based regions of DSA minimum intensity projections (MinIPs) for carotid terminus detection and occlusion classification. Warman et al. \cite{warmanDeepLearningMethod2024a} used an ensemble method, in which a DSA frame was evaluated by a ResNet-18-based detection model to predict if it had a large vessel occlusion (LVO), in which case the frame was then fed into a localization model based on a YOLOv5 backbone to localize the LVO with a bounding box. These approaches, however, treat DSA as a collection of independent images or collapse them into a single view using MinIP, ignoring the temporal information contained in the images.
Leveraging the temporal nature of DSA imaging, Mittmann et al. \cite{mittmannDeepLearningbasedClassification2022} demonstrated that a thrombus yes-no-classification can be achieved by combining EfficientNet as a feature extractor with a bidirectionally configured long-short term memory (LSTM) or a gated recurrent unit (GRU) to establish spatio-temporal relations between the individual images of the DSA sequence. Kelly et al. \cite{kellyDEEPMOVEMENTDeep2023} evaluated the performance of several DL models, including a 2D CNN Xception (using only single DSA frames), a stacked 2D CNN stacked-Xception (using multiple DSA frames), a 2D vision transformer (ViT), and a 3D-Inception CNN, the latter capturing full spatial and temporal resolution and achieving high sensitivity in occlusion detection using pre- and post-thrombectomy DSA video sequences. These studies confirmed that temporal context is critical for occlusion classification, particularly for distal occlusions where small thrombi may only become apparent through sequential frame analysis. Methods that explicitly localize the occlusion in the DSA are still underexplored. We hypothesized that occlusion detection on DSA is a task that can benefit from integration of spatio-temporal information, due to the inherently temporal aspect of occlusion visualization on DSA. 

Building upon these advances in spatio-temporal DSA analysis, we develop a method that explicitly localizes the occlusion using a bounding box. We introduce OccluNet, a DL pipeline that combines YOLOX \cite{geYOLOXExceedingYOLO2021}— a high-performance single-stage object detector— with transformer-based temporal modules. Our proposed architecture compares two attention mechanisms for temporal learning: (1) a self-attention module that processes YOLO-extracted features across frames to model long-range contrast flow dependencies, inspired by transformer-based temporal learning approaches \cite{hanCoronaryArteryStenosis2023,xieDSCADigitalSubtraction2024a,suspatiotemporalDeepLearning2022a,suCAVECerebralArtery2024,xieDSNetspatiotemporalConsistency2024}; and (2) a divided space-time attention mechanism adapted from the TimeSFormer architecture \cite{bertasiusSpaceTimeAttentionAll2021}. We compare both approaches with a baseline model trained on MinIPs, demonstrating superior performance of the spatio-temporal approach. 

\begin{figure}[h!]
        \centering
        \includegraphics[width = \textwidth]{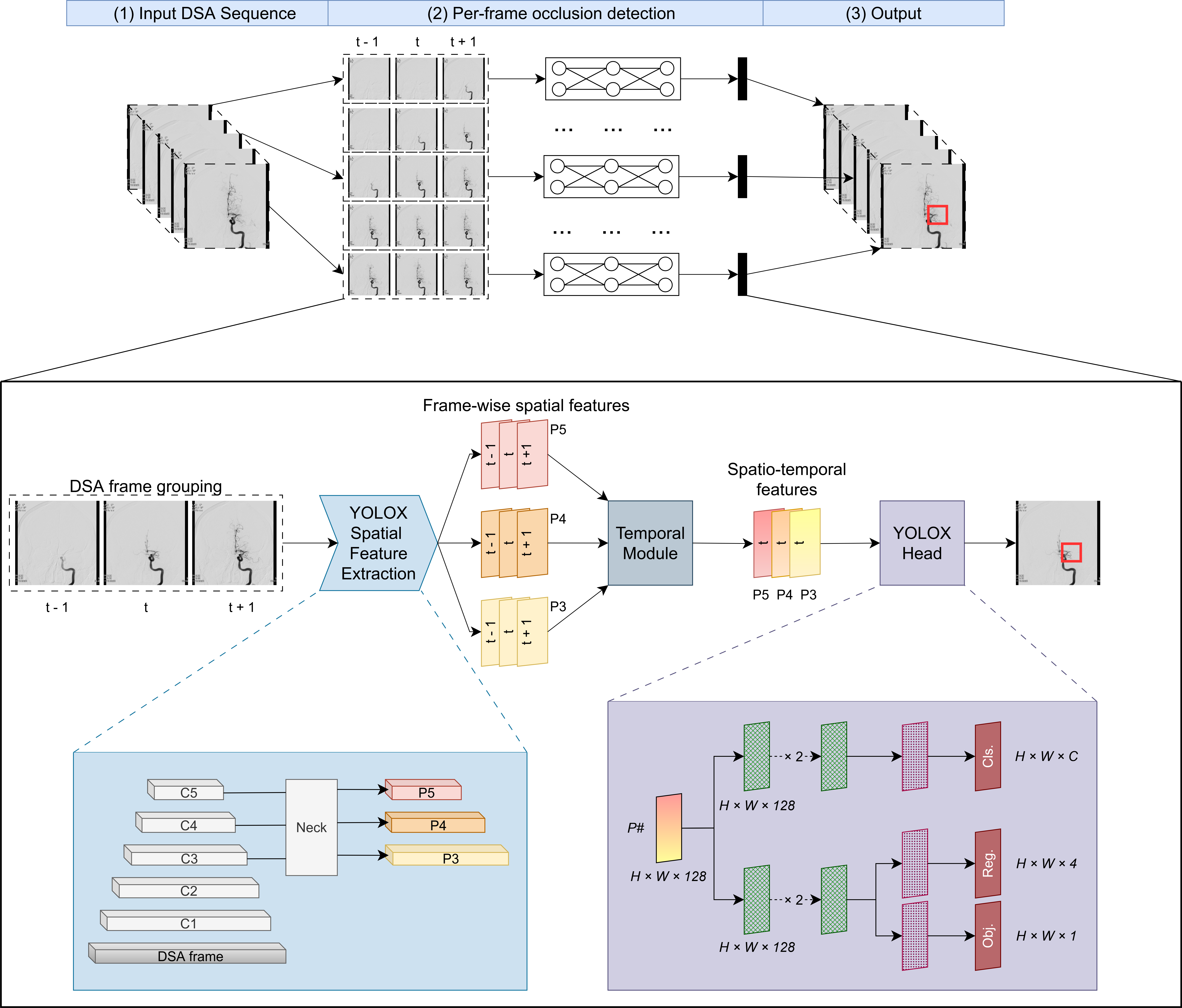}
        \caption{Overview of OccluNet. Given an input DSA sequence (1), occlusions are first detected on each current frame (timepoint $t$) by taking neighbouring frames ($t \pm 1$) as context (2), and these per-frame detections form the final output (3). The model consists of a PAFPN feature extraction module, a temporal module and a YOLOX head. For end-to-end per-frame occlusion detection, first, frame-wise spatial features are extracted using the YOLOX spatial feature extraction module. The temporal module applies (temporal or divided space-time) attention to each frame-wise spatial feature level. The output of the temporal module constitutes enhanced spatio-temporal features at each feature level for the current frame, with incorporated context from the neighbouring frames. Finally, the YOLOX decoupled head performs the occlusion detection for each level of spatio-temporal features.}
        \label{fig:custom_model}
\end{figure}

\section{Methods}
\subsubsection{YOLOv11 Baseline}
To establish baseline results, we used the X variant of the \text{YOLOv11} model with 56.9 million parameters, the latest iteration of the You Only Look Once (YOLO) series of real-time object detection models. The YOLOv11 model is commonly used across multiple vision tasks, including detection, segmentation, and pose estimation \cite{UltralyticsYOLO11Ultralytics}. YOLOv11 predicts three outputs, bounding box locations, class probability ($P(\text{class}_i)$) and objectness probability ($P(\text{object})$). From these probabilities indicating whether a bounding box is an object, and which class that object is a member of, a confidence score can be calculated per bounding box, $\text{Confidence}_{\text{class}_i} = P(\text{object}) \times P(\text{class}_i)$. 

\subsubsection{OccluNet}
Two custom architectures ($\text{OccluNet1}$ and $\text{OccluNet2}$) were developed, based on different temporal modules. For both models, the backbone is a pretrained CSPDarknet, the same backbone as used in the S variant of YOLOX with nine million parameters. To facilitate spatio-temporal processing, the current and neighboring frames ($t-1,t,t+1$) were passed as input channels. The backbone extracts multi-scale spatial features (C3, C4, C5) from each input frame stack. These frame-wise spatial features are then passed through a Path-Aggregation Feature Pyramid (PAFPN) neck module, as used by YOLOX, followed by a temporal module. There are two temporal approaches:
\begin{itemize}
    \item  $\text{OccluNet1}$ uses Temporal Attention, a Transformer-based attention across frames at each spatial location, using learned positional encoding.
    \item  $\text{OccluNet2}$ uses Divided Space-Time Attention, which alternates between spatial and temporal attention layers, separating spatial ("where") and temporal ("when") dynamics for occlusion detection.
\end{itemize}
After temporal processing for the current frame group, we are left with spatio-temporal features at three levels (P3, P4, P5). These features are then processed by the YOLOX head module, containing the same three output models as the baseline. To enforce temporal consistency, we applied a trajectory-based optimization \cite{suspatiotemporalDeepLearning2022a}, linking detections across frames if their centers lay within a $15\text{px}$ radius. Each trajectory’s score was the sum of its confidence scores, multiplied by duration (number of frames) to reward temporally persistent predictions. The highest-scoring trajectory per sequence was selected, suppressing inconsistent false positives. See Figure \ref{fig:custom_model} for an overview of OccluNet. 

\section{Experiments \& Results}

\subsection{Data}
The \anonymize{MR CLEAN} Registry is a \anonymize{Dutch multi-center registry} of acute ischemic stroke patients who underwent EVT between \anonymize{March 2014 and December 2018}. The DSA sequences included in this dataset were acquired using various imaging systems. We included one lateral and one anteroposterior pre-EVT DSA per patient, ensuring visibility of the occlusion. These sequences are of size $1024 \times 1024$ pixels, containing between 3 and 76 frames. The temporal resolution varied from 1 to 4 frames per second, while the spatial resolution was approximately $0.4\text{mm}\times0.4\text{mm}$ per pixel. 
From the \anonymize{MR CLEAN} Registry, $n=1061$ pre-EVT DSA DICOM sequences were selected from $809$ patients, containing $19144$ DSA frames in total. Six different occlusion location categories were considered, M1 ($n=450$), M2 ($n=238$), Intracranial ICA ($n=192$), Extracranial ICA ($n=28$), Unknown ($n=61$), and No occlusion ($n=92$). 


\subsubsection{Data Preprocessing \& Annotation}
DSA sequences were normalized and resized to target dimensions of $640 \times 640$ pixels using bicubic interpolation. Resizing was done due to computational constraints. During training, random left-right flipping was implemented at the sequence level, with a probability of $50\%$. 
A custom annotation tool was developed using MeVisLab, a platform for medical image processing and visualization (\url{https://www.mevislab.de/}). Using this tool, we annotated occlusion bounding box ($40\text{px} \times 40\text{px}$) coordinates, occluded vessel type (Extracranial ICA, Intracranial ICA, M1, M2, Unknown, and No occlusion), and the frame interval when the occlusion was visible, if present. All cases were annotated by \anonymize{AK} and reviewed by \anonymize{MvdS, FtN}. Challenging cases, such as those with ambiguous occlusions or multiple vessel involvement, were discussed with an experienced neurointerventionalist (\anonymize{SC}) for verification. 

\subsection{Evaluation Metrics}
Given the small size of the resampled occlusion bounding boxes ($\text{25px} \times 25$px) relative to the resampled DSA frame dimensions ($640 \times 640$ pixels), we adopted lenient evaluation criteria. In clinical practice, the exact location of the occlusion can be ambiguous, justifying our relatively lenient criteria. Predictions were assessed at the DSA sequence level. Initial per-frame outputs included bounding boxes, confidence scores, and class labels. A detection was considered true positive ($TP$) if the distance between centers of the highest-scoring trajectory and the ground truth bounding box for the DSA sequence is at most $25\text{px}$, with a minimum confidence score of 0.01, and a correct class label. False positives ($FP$) occurred when the distance exceeded 25 px or the class was incorrect, while false negatives ($FN$) represent undetected ground truth occlusions. 

\subsection{Implementation}
The YOLOv11 baseline model was trained for 200 epochs with a batch size of 16. The remaining arguments used the default values recommended for training YOLOv11 \cite{ModelTrainingUltralytics}. 
OccluNet was implemented in Python using PyTorch \cite{paszkePyTorchImperativeStyle2019} and the MMDetection library \cite{mmdetection}. The models were trained on an NVIDIA A40 with 46 GB of memory. OccluNet variants were trained for 20 epochs. 
The batch size was set to four and one for training and validation, respectively. The input sequence length was set to 3. We employed standard YOLOX training protocols, using stochastic gradient descent (SGD) with Nesterov momentum ($0.9$) as the optimizer. The base learning rate was set to $0.005$, with a two-phase learning schedule- a quadratic warm-up over the first two epochs for stable initialization, followed by cosine annealing that gradually reduces the learning rate to 0.1\% of its initial value over the remaining 18 epochs. The final two epochs used a fixed low learning rate for fine-tuning. Weight decay was set to $5\mathrm{e}^{-4}$. 

\renewcommand{\arraystretch}{1.2}
\begin{table*}[t]
\centering
\begin{tabular}{llllllll}
\hline
\textbf{Exp.} & \textbf{Dataset} & \textbf{Model Type} & \textbf{Class} & \textbf{Samples} & \textbf{Instances} & \textbf{P (\%)} & \textbf{R (\%)} \\
\hline
\multirow{6}{*}{1} 
& \multirow{6}{*}{MinIP\textsubscript{M}} & \multirow{6}{*}{YOLOv11} & all & 214 & 195 & 13.4 & 18.4 \\ 
& & & M1 & 87 & 87 & 30.0 & 39.1 \\ 
& & & M2 & 50 & 50 & 7.56 & 16.0 \\ 
& & & Intracranial ICA & 38 & 38 & 29.6 & 36.8 \\ 
& & & Extracranial ICA & 4 & 4 & 0.0 & 0.0 \\ 
& & & Unknown & 16 & 16 & 0.0 & 0.0 \\ 
\hline
2 & MinIP\textsubscript{S} & YOLOv11 & all & 214 & 195 & 44.5 & 22.6 \\
\hline
\multirow{6}{*}{3} 
& \multirow{6}{*}{DSA} & \multirow{6}{*}{OccluNet1} & all & 214 & 195 & 89.02 & 74.87 \\ 
& & & M1 & 87 & 87 & 84.81 & 77.01 \\ 
& & & M2 & 50 & 50 & 52.38 & 44.0 \\ 
& & & Intracranial ICA & 38 & 38 & 93.75 & 78.94 \\ 
& & & Extracranial ICA & 4 & 4 & 100.0 & 100.0 \\ 
& & & Unknown & 16 & 16 & 45.45 & 31.25 \\ 
\hline
\multirow{6}{*}{4} 
& \multirow{6}{*}{DSA} & \multirow{6}{*}{OccluNet2} & all & 214 & 195 & 87.97 & 71.28 \\ 
& & & M1 & 87 & 87 & 84.61 & 75.86 \\ 
& & & M2 & 50 & 50 & 53.65 & 44.0 \\ 
& & & Intracranial ICA & 38 & 38 & 90.0 & 71.05 \\ 
& & & Extracranial ICA & 4 & 4 & 100.0 & 75.0 \\ 
& & & Unknown & 16 & 16 & 33.33 & 18.75 \\ 
\hline
\end{tabular}
\caption{Performance comparison of baseline (YOLOv11) and OccluNet models across different datasets and occlusion classes. The MinIP\textsubscript{M} and MinIP\textsubscript{S} experiments represent multi-class and single-class detection on MinIP images of complete DSA sequences, while OccluNet1 and OccluNet2 were evaluated on full DSA sequences. Metrics are reported for both aggregate performance ("all") and individual occlusion classes. Exp.: Experiment, P: Precision, R: Recall.}
\label{tab:results}
\end{table*}

\subsection{Results}
In total, we conducted four experiments. Experiments 1-2 assessed the YOLOv11 baseline model performance on MinIPs. For experiment 1, we trained the baseline model on MinIP images to detect individual occlusion classes. Experiment 2 assesses baseline performance on MinIPs when detecting a single "Occlusion" class, i.e., treating all occlusions as one class. Experiments 3-4 investigate OccluNet configurations on full DSA sequences, with Experiment 3 assessing the performance of the temporal attention variant (OccluNet1) in detecting single-class occlusions on DSA, whereas Experiment 4 assesses the divided space-time attention variant (OccluNet2). 
The YOLOv11 baseline performance is detailed in Table \ref{tab:results}. When processing MinIPs (Exp. 1 and 2), multi-class detection (Exp. 1) achieved an overall precision of 13.4\%, with varying performance on the subclasses ranging from 30\% for the M1 occlusions to 0.0\% for Extracranial and Unknown occlusions. Recall ranged from 39.1\% to 0.0\%. Single-class occlusion detection (Exp. 2) shows a precision of 44.5\% and a recall of 22.6\%.

\begin{figure}[t]
        \centering
        \includegraphics[width = \textwidth]{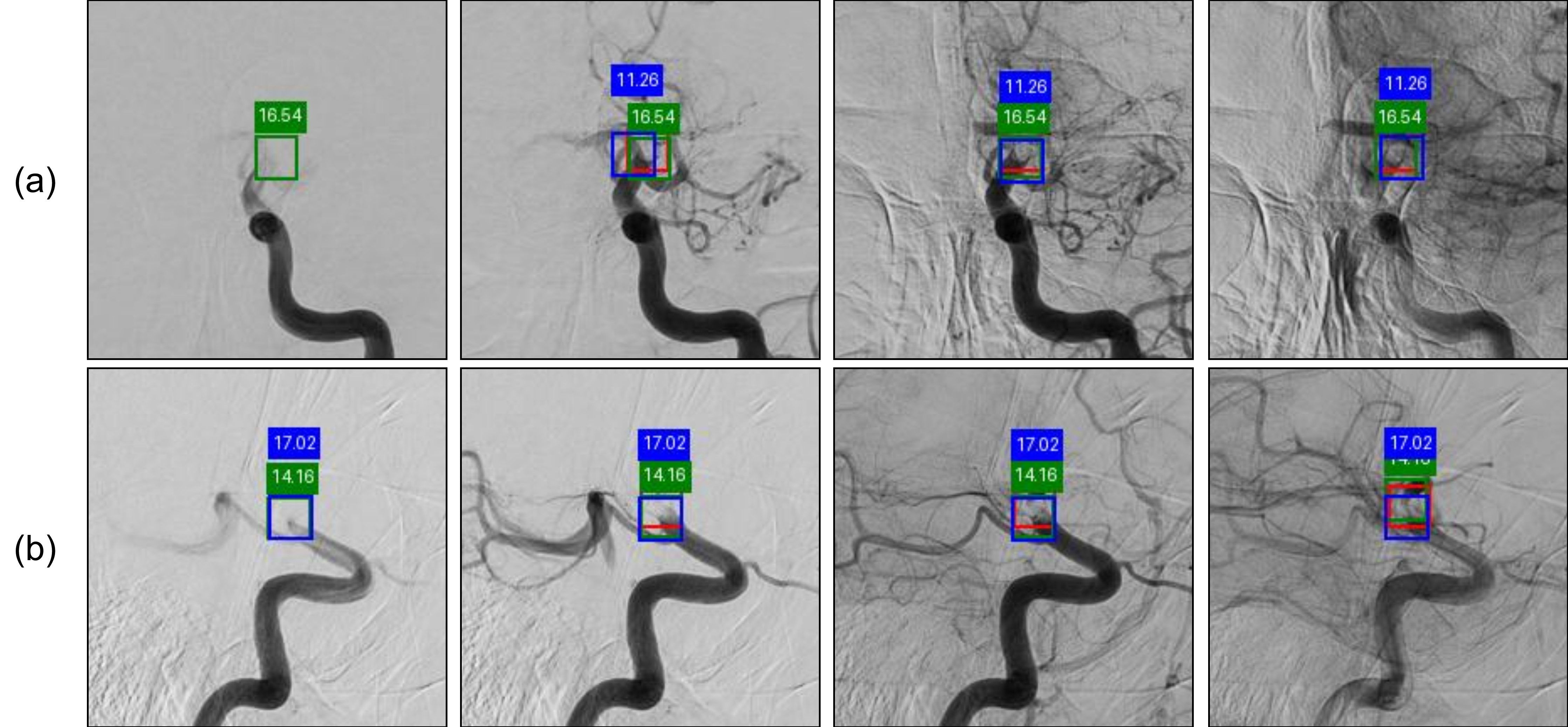}
        \caption{Visualization of occlusion detection results after trajectory-based optimization to enforce temporal consistency. (a) represents an anteroposterior DSA sequence, and (b) a lateral DSA sequence from the same patient. The red bounding boxes indicate the ground truth annotations, while the green and blue bounding boxes represent predicted occlusions by OccluNet1 and OccluNet2, respectively. The numbers indicate the detection confidence scores.}
        \label{fig:preds}
\end{figure}

OccluNet demonstrated distinct performance characteristics, as detailed in Table \ref{tab:results}. On clinical DSA data (Exp. 3 and 4), both temporal attention variants achieved strong performance, with the temporal attention module (Exp. 3- $\text{OccluNet1}$) showing an overall precision and recall of 89.02\% and 74.87\%, respectively, with a particularly strong performance on Intracranial ICA (93.75\% precision, 78.94\% recall) and perfect detection of Extracranial ICA occlusions (100\% precision, 100\% recall). The divided space-time variant (Exp. 4- $\text{OccluNet2}$) reached similar values of overall precision and recall of 87.97\% and 71.28\%, respectively. McNemar's test confirmed these improvements were statistically significant when comparing both OccluNet variants to the YOLOv11 baseline (MinIP\textsubscript{S} vs OccluNet1: $p < 0.001$, MinIP\textsubscript{S} vs OccluNet2: $p < 0.001$), demonstrating the importance of temporal information. McNemar's test showed no significant differences between OccluNet1 and OccluNet2 performance ($p = 0.60$). Example of predictions made by both OccluNet configurations on clinical DSA after implementing temporal consistency can be seen in Fig. \ref{fig:preds}.


\section{Discussion}


We presented and assessed OccluNet, a novel spatio-temporal deep learning framework that integrates a YOLOX-based object detector with temporal attention mechanisms to exploit the dynamic flow information inherent in DSA sequences. We compared the performance of OccluNet to a YOLOv11 baseline model, showing a significant ($p < 0.001$) performance improvement when incorporating spatio-temporal information.
The YOLOv11 baseline model is unable to incorporate temporal features, as it is trained on MinIP images (Exp. 1 and 2). OccluNet, by contrast, is able to assess temporal information by processing an entire DSA sequence during a single inference, either attending across time at each spatial location (OccluNet1) or by sequentially attending to temporal and spatial information (OccluNet2). The fact that both OccluNet variants significantly outperform the baseline models indicates that the integration of temporal information is beneficial. These findings are in line with human behavior, as neurointerventionalists often "scroll through" the images to detect distinctive contrast flow patterns indicative of occlusions.
OccluNet was trained for 20 epochs due to computational constraints, whereas the baseline was less demanding, allowing for a training over 200 epochs. Despite this imbalance, OccluNet outperformed the baseline models, highlighting the importance of temporal dynamics in DSA processing.
The YOLOv11 baseline models showed increased performance when trained to detect any occlusions (single-class detection) compared to the models trained to detect multiple occlusions (multi-class detection) and differentiate between them. This behavior is expected, as it is easier for models to learn to detect a single class than to further differentiate between them.
When comparing OccluNet1 and OccluNet2, we did not see statistically significant performance differences between the models (p-value= $0.6)$, indicating that both attention types are able to achieve similar performance. From this, we conclude that it is not relevant how exactly temporal information is integrated in an occlusion detection pipeline, as long as it is integrated in some form.

Methodologically, OccluNet uses the YOLOX-S variant as its backbone. This is a relatively simple backbone which could be replaced with more complex variants, such as YOLOX-L, to make the model more powerful, potentially improving performance. Due to computational restrictions, we were unable to evaluate this. 
Integrating anatomical priors (e.g., vessel segmentation masks) or multimodal inputs (e.g., adding CTA) may reduce false positives by constraining detections to plausible regions. Finally, more diverse datasets with standardized annotations for distal, rare occlusion types and multi-occlusion cases will be important for improving generalization.


OccluNet outperforms previous DSA occlusion detection models based on reported metrics. Future work should assess the applicability of OccluNet in a clinical setting. The detection of distal occlusions is the most clinically relevant, as these occlusions are easier to miss on DSA. As such, clinical evaluation of this model should focus on the added benefit in the detection of distal occlusions (M3, M4) in pre-EVT DSA as well as fragmented thrombi in post-EVT assessment. The relatively low prevalence of these distal occlusion types in DSA datasets presents a challenge.

\section{Conclusion}
This work presents OccluNet, a spatio-temporal deep learning framework for automated occlusion detection in DSA sequences, with the potential to assist radiologists during EVT procedures for AIS. 
OccluNet significantly outperforms baseline spatial models, suggesting a role for spatio-temporal processing in automated occlusion detection on DSA. This work lays the foundation for tools that improve clinical workflows during endovascular stroke interventions.

\subsubsection{Acknowledgments.} We thank the \href{https://www.linkedin.com/company/icai-stroke-lab/}{ICAI Stroke Lab} collaborators for their contributions. This publication is part of the project ROBUST: Trustworthy AI-based Systems for Sustainable Growth with project number KICH3.LTP.20.006, which is (partly) financed by the Dutch Research Council (NWO), Philips Medical Systems Nederland B.V., and the Dutch Ministry of Economic Affairs and Climate Policy (EZK) under the program LTP KIC 2020-2023.

\subsubsection{Disclosure of Interests.} The authors have no competing interests to declare that are relevant to the content of this article.

%
%
\newpage
\bibliographystyle{splncs04}
\bibliography{Bibliography}
%




\end{document}